\documentclass[10pt,twocolumn,letterpaper]{article}
\usepackage{cvpr}
\usepackage{graphicx}
\usepackage{comment}
\usepackage{amsmath,amssymb} 
\usepackage{color}
\usepackage{times}
\usepackage{epsfig}
\usepackage{diagbox}
\usepackage{afterpage}
\usepackage{makecell}
\usepackage{lineno}
\usepackage{amsfonts}       
\usepackage{amsthm}   

\usepackage{bm}
\usepackage{subfigure}
\usepackage{graphicx}
\usepackage{algorithm}
\usepackage{algorithmic}

\usepackage{amssymb}
\usepackage{caption}

\newcommand{\be}{\begin{eqnarray}}
\newcommand{\ee}{\end{eqnarray}}
\newcommand{\beq}{\begin{equation}\begin{aligned}}
\newcommand{\eeq}{\end{aligned}\end{equation}}
\newcommand{\beqn}{\begin{equation*}\begin{aligned}}
\newcommand{\eeqn}{\end{aligned}\end{equation*}}
\newcommand{\ben}{\begin{eqnarray*}}
\newcommand{\een}{\end{eqnarray*}}
\newcommand{\bena}{\begin{eqnarray*}\begin{aligned}}
\newcommand{\eena}{\end{aligned}\end{eqnarray*}}
\newcommand{\bea}{\begin{eqnarray}\begin{aligned}}
\newcommand{\eea}{\end{aligned}\end{eqnarray}}
\usepackage{algorithm}
\usepackage{algorithmic}

\usepackage{multirow}
\usepackage{booktabs}       

\usepackage[pagebackref=true,breaklinks=true,letterpaper=true,colorlinks,bookmarks=false]{hyperref}

\usepackage{authblk}
\cvprfinalcopy
\ifcvprfinal\fi
\begin{document}

\title{Adversarial YOLO: Defense Human Detection Patch Attacks\\ via Detecting Adversarial Patches} 
\author{Nan Ji$^\ddagger$, YanFei Feng$^\ddagger$, Haidong Xie, Xueshuang Xiang$^\ast$, Naijin Liu}
\affil{\normalsize Qian Xuesen Laboratory of Space Technology\\ China Academy of Space Technology}
\date{}

\maketitle

\let\thefootnote\relax\footnotetext{\noindent$^\ddagger$~These authors contributed equally.}
\let\thefootnote\relax\footnotetext{\noindent$^\ast$~Corresponding author: xiangxueshuang@qxslab.cn}

\begin{abstract}
The security of object detection systems has attracted increasing attention, especially when facing adversarial patch attacks. Since patch attacks change the pixels in a restricted area on objects, they are easy to implement in the physical world, especially for attacking human detection systems. The existing defenses against patch attacks are mostly applied for image classification problems and have difficulty resisting human detection attacks. Towards this critical issue, we propose an efficient and effective plug-in defense component on the YOLO detection system, which we name Ad-YOLO. The main idea is to add a patch class on the YOLO architecture, which has a negligible inference increment. Thus, Ad-YOLO is expected to directly detect both the objects of interest and adversarial patches. To the best of our knowledge, our approach is the first defense strategy against human detection attacks.

We investigate Ad-YOLO's performance on the YOLOv2 baseline. To improve the ability of Ad-YOLO to detect variety patches, we first use an adversarial training process to develop a patch dataset based on the Inria dataset, which we name Inria-Patch.
Then, we train Ad-YOLO by a combination of Pascal VOC, Inria, and Inria-Patch datasets. With a slight drop of $0.70\%$ mAP on VOC 2007 test set, Ad-YOLO achieves $80.31\%$ AP of persons, which highly outperforms $33.93\%$ AP for YOLOv2 when facing white-box patch attacks. Furthermore, compared with YOLOv2, the results facing a physical-world attack are also included to demonstrate Ad-YOLO's excellent generalization ability.
\end{abstract}

\section{Introduction}
Patch attacks have been recognized as a very practical means to threaten computer vision systems. Unlike the traditional strategy, patch attacks only change the pixels in the restricted area and are not imperceptible, which is close to the effect of graffiti or stickers in form. Therefore, it is more realistic in physical-world attacks. At present, many carefully crafted adversarial patches have achieved remarkable results in the fields of image classification \cite{kurakin2017adversarial}, face recognition \cite{sharif2018adversarial} and object detection \cite{eykholt2018robust,thys2019fooling,lee2019on}.

Compared with rich work on patch attacks, considerable attention for patch defenses has been devoted to image classification and cannot be transferred to object detection due to the large computational cost \cite{wu2020defending,raghunathan2018certified} or the lack of prior knowledge about the patch \cite{xiang2020patchguard}. In addition, there are some image preprocessing methods for mitigating adversarial noise for patch attacks, such as image denoising DW \cite{hayes2018on}, LGS \cite{naseer2019local} and partial occlusion MR \cite{mccoyd2020minority}, but the drawback is the accuracy decrease on the original examples. Besides, most of these defenses are designed to resist patch attacks in the digital space and are ineffective to defend against physical-world attacks. 

This paper focuses on patch defenses for objection detection, especially human detection systems. We first conjecture that a qualified defense should have the following characteristics:

\begin{itemize}	
	\item \textbf{Timeliness}: the method shows good real-time performance against patch attacks.   
	\item \textbf{Detectability}: the method cannot affect the detector to recognize persons without attacks.
	\item \textbf{Robustness}: the method is resilient to white-box patch attacks and shows generalization to different scenarios and persons.	
\end{itemize}

To achieve the above three properties, we establish an efficient and effective plug-in defense component on the YOLO detection system, which we name Ad-YOLO, that enhances the resilience against patch attacks. Specifically, observing that the object detection model YOLOv2 has a high ability for detecting two types of objects with overlapping regions, we add a new category of adversarial patch to YOLOv2. Therefore, we first develop a patch dataset by an adversarial training process based on the Inria dataset, which we name Inria-Patch, and train Ad-YOLO by a combination of Pascal VOC, Inria, and Inria-Patch datasets. With our defense method, Ad-YOLO is expected to directly detect adversarial patches and persons at the same time when facing with a physical-world attack.

\begin{figure}[htbp]				
	\begin{center}
		\includegraphics[width=0.45\linewidth]{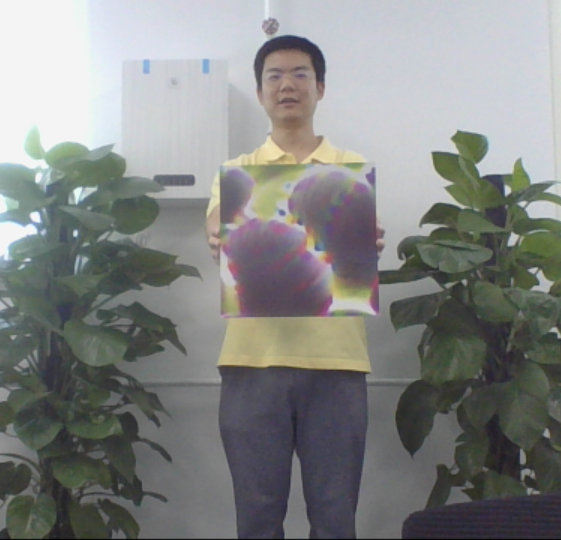}\hspace{0.1 in}
		\includegraphics[width=0.45\linewidth]{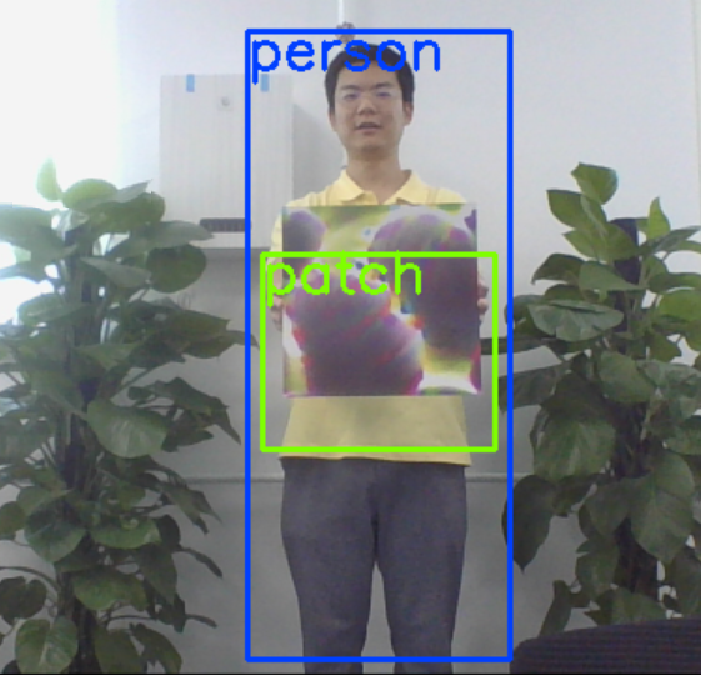}
		\caption*{(a) YOLOv2\hspace{0.9 in}(b) Ad-YOLO}
		\caption{A demonstration of the YOLOv2 and Ad-YOLO when facing with a physical-world attack. (a) YOLOv2: a person with an adversarial patch is ignored. (b) Ad-YOLO: the person and the adversarial patch are both detected.}
		\label{Fig:demonstration}	
	\end{center}		
\end{figure}

We demonstrate the performance of Ad-YOLO in detectability and robustness with a series of experiments. The results indicate that Ad-YOLO can only reduce the mAP on VOC 2007 test set from $73.07\%$ to $72.35\%$, the AP of persons on Inria test set from $88.13\%$ to $86.91\%$, but exhibit $83.91\%$ AP of patch on Inria-Patch dataset. This means that Ad-YOLO can both detect persons and adversarial patches with the same level and remarkable precision. In addition, when facing white-box patch attacks, Ad-YOLO can also achieve $80.31\%$ AP of persons, which highly outperforms $33.93\%$ AP for YOLOv2. Furthermore, compared with YOLOv2, the results facing a physical-world attack are also included to demonstrate Ad-YOLO's excellent generalization ability. Figure \ref{Fig:demonstration} demonstrates the performance of YOLOv2 and Ad-YOLO when facing with a physical-world attack in a human detection system.

Our main contributions can be summarized as follows:
\begin{itemize}	
	\item We propose a defense model called Ad-YOLO against patch attacks in a human detection system, which adds the patch category to the YOLOv2 model and maintains high detection capability both on persons and adversarial patches.
	\item We develop a patch dataset named Inria-Patch containing 200 adversarial patches with diversity and adversariality.
	\item We provide a comprehensive analysis for generalization of Ad-YOLO to different scenarios and persons. In addition, We examine the performance of Ad-YOLO against white-box patch attacks and physical-world attacks.
\end{itemize}

\section{Related work}
Most published defenses against patch attacks focus on image classification. \cite{raghunathan2018certified} proposed the first certified defense against patch attacks based on interval bound propagation and two training strategies to speed up the approach while trading-off efficiency and robustness. In \cite{xiang2020patchguard}, PatchGuard was proposed to use convolutional networks with small receptive fields that impose a bound on the number of features corrupted by an adversarial patch. However, the premise is to have a priori knowledge of the size of the patch, which is unrealistic  in physical-world attacks. In \cite{wu2020defending}, rectangular occlusion attacks (ROA), coupled with adversarial training, were proposed to achieve high robustness against several prominent examples of physical-world attacks, but it is too computationally expensive for object detection systems.

In addition, some work is based on preprocessing input images to mitigate adversarial noise but cannot be transplanted to object detection. In \cite{hayes2018on}, the proposed digital watermarking (DW) exploited important pixels that affect the gradient of the loss and mask them out in the image. In \cite{naseer2019local}, local gradient smoothing (LGS) was based on the empirical observation that the image gradients tend to be large within these adversarial patches. However, the above two methods significantly reduce the accuracy of unattacked images and are easily broken by white-box attacks. The minority report (MR) \cite{mccoyd2020minority} proposed a defense against patch attacks based on partially occluding the image around each candidate patch location, but this defense only detects the presence of an attack, rather than recovering the correct prediction.

As far as we know, patch defenses have not been rigorously explored in object detection systems. In \cite{saha2020role}, Saha et al. proposed a defense algorithm by regularizing the model to limit the influence of image regions outside the bounding boxes of the detected objects. This method is only applicable when the patch does not occlude the object in digital space, which is different from the scenario discussed in this paper. For defenses against physical-world attacks, Wu, Tong et al. \cite{wu2020defending} made an attempt on the image classification. Due to the huge amount of computational cost, it is difficult to incorporate into the object detection.

\section{Methodology}
\begin{figure*}[t]
	\centering
	\includegraphics[width=0.9\linewidth]  {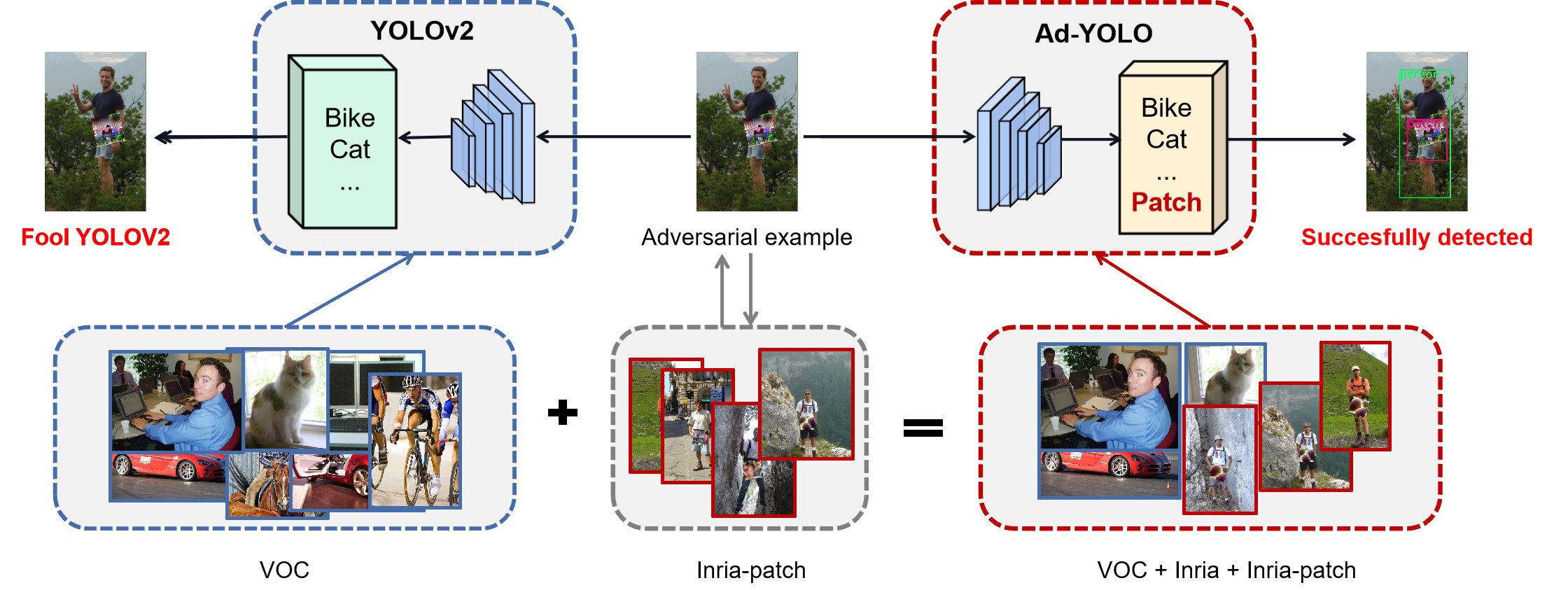}
	\caption{ Overview of the Ad-YOLO defense framework. By adding the "Patch" category to the last layer of the YOLOv2 network while maintaining the structure of other layers, Ad-YOLO can simultaneously identify the adversarial patch and recover the correct prediction of the original object. Pascal VOC, Inria together with the newly constructed dataset Inria-Patch form the training set of Ad-YOLO.}
	\label{Ad-YOLO}
\end{figure*}

\subsection{Attack Method}
Generally speaking, there are currently two main solutions to patch attacks in human detection systems. The difference is whether we put the patch in the background \cite{thys2019fooling} or on the objects of interest \cite{lee2019on}. 
Considering that environmental factors are difficult to control in actual scenarios, the method of placing patches on the object is more in line with the needs of the application. Therefore, we use the method proposed in \cite{thys2019fooling} to attack the object detection model YOLOv2. 

First, we review the one-stage detection system YOLOv2 architecture. YOLOv2 divides a single frame image into multiple grids, and predicts bounding boxes for each grid cell. Assuming that there are $n$ classes to be identified, each bounding box can be written as an $(n+5)$ dimensional vector, as shown below:
\begin{equation}\label{anchor box}
(\hat{X}, \hat{Y}, W, H, P_{obj}, P_{cls_{1}}, ..., P_{cls_{n}}).
\end{equation}
where $\hat{X}$ and $\hat{Y}$ represent the center of the bounding box. $W$ and $H$ represent the width and height predicted relative to the whole image. $P_{obj}$ is the confidence score that represents the intersection over union (IOU) between the predicted box and any ground truth box, and $n$ conditional class probabilities $P_{cls_{i}}$ are conditioned on the grid cell containing an object.

To attack the above model, the method \cite{thys2019fooling} considers the following mathematical formulation of finding an adversarial patch:
\begin{equation}
{arg}\min_\delta {E}_{(x,y)\sim\mathcal{D},t\sim T}[J(A(\delta,x,t),y)],
\label{generate_patch}
\end{equation}
where $\mathcal{D}$ is a distribution over samples, $T$ ia s distribution over patch transformations, and $A$ is a patch application function that transforms the patch $\delta$ with $t$ and applies the result to the image $x$ by masking the appropriate pixels. The loss function $J(A(\delta,x,t),y)$ consists of three parts, as shown below:
\begin{equation}
J=\alpha J_{nps}+\beta J_{tv}+J_{obj}.
\end{equation}
The first item here is non-printable fraction denoted as $J_{nps}$ that represents how well the colours in the patch can be represented by a common printer. The second item, $J_{tv}$, is the total variation value of the image, which makes the patch color transition smooth and more natural. The third $J_{obj}$ represents the probability that the detector will detect a person, and we want this value to be as small as possible. We assume that people happen to be $cls_i$, then, there are three choices for $J_{obj}$ , namely, $P_{obj}$, $P_{cls_{i}}$ and $P_{obj} P_{cls_{i}}$. In \cite{thys2019fooling}, experiments are conducted on these three loss functions, and it is found that $P_{obj}$ has the strongest attack effect. In this paper, we set $J_{obj}=P_{obj}$ to generate patches.

The attack proposed in \cite{thys2019fooling} can generate a rectangular patch, which can be printed out and taped to people's chest during the attack. Thus, the accuracy of the detector in identifying a person can be greatly reduced, making the detector unable to identify the presence of a person, as shown in Figure \ref{Fig:demonstration}(a). This type of attack in the physical world is inexpensive and highly threatening, so it is imperative to formulate corresponding defense strategies. 
\subsection{Adversarial YOLO}
Motivated by the fact that object detection model YOLOv2 can recognize overlapping objects, in this paper, we develop an Ad-YOLO model that can recognize adversarial patches as a new category and restore the original object information. Additionally, to achieve better defense, we construct a new dataset, Inria-Patch, to assist in training the Ad-YOLO model.

\subsubsection{Framework}
Figure \ref{Ad-YOLO} provides an overview of our defense framework. From the structure of the neural network, Ad-YOLO retains all layers of YOLOv2 except for the last layer by adding a ``patch'' category. Therefore, Ad-YOLO is expected to recognize the objects and patches from the input image at the same time, see the right part of Figure \ref{Ad-YOLO}.

As we introduced, YOLOv2 divides an input image $x$ into $P^2$ preselected areas, with each area predicts $B$ anchor boxes. Each anchor box is a vector with dimension $(5+n)$, as shown in (\ref{anchor box}). We then arrange the anchor boxes in order, and thus each preselected area would output a vector with dimension $B(5 + n)$. Ultimately, the output of YOLOv2 can be expressed as a tensor of $P^2 B (5 + n)$. 
For Ad-YOLO, we only add the patch category in the last layer of YOLOv2, so the length of each anchor box vector becomes $(5+n+1)$, and the corresponding final output becomes a tensor with dimension $P^2 B (5 + n + 1)$.

We write $L_{\theta}(x, y)$ as the loss function of Ad-YOLO, which inherits the full form of YOLOv2 loss function and add the loss for the conditional patch class probability. We assume that the adversarial patches we generate to train Ad-YOLO are $\Phi = \{\delta\}$. Then, our method can be expressed as:
\begin{equation}\label{eq-Ad-YOLO}
\min_{\theta} {E}_{(x^{\prime},y^{\prime})\sim\mathcal{D^{\prime}}}L_{\theta}(x^{\prime}, y^{\prime}),
\end{equation}
where $D^{\prime} $ is a distribution over samples, $x^{\prime}$ and $y^{\prime}$ as shown below:
\begin{equation}
(x^{\prime},y^{\prime})=\left\{
\begin{array}{cl}
(x,y), &  x\in \text{Pascal VOC}\cup\text{Inria} \\
(A(\delta, x, t),\hat{y}), &  x\in \text{Inria},\delta\in\,\Phi
\end{array} \right.
\end{equation}
$A(\delta, x, t)$ is a patch application function mentioned in (\ref{generate_patch}), $\hat{y}$ is the new label formed after the addition of a patch category to the ground true label $y$ of input image $x$.

From the above description, it can be concluded that Ad-YOLO has two advantages.
On the one hand, Ad-YOLO inherits the reference effectiveness of YOLOv2 and satisfies the requirement on \textbf{timeliness} to defend against a physical-world attacks.
On the other hand, Ad-YOLO can also be regarded as a model for detecting attacks in a sense.

\subsubsection{Inria-Patch Dataset}
\begin{figure}
	\centering
	\includegraphics[width=0.9\linewidth]  {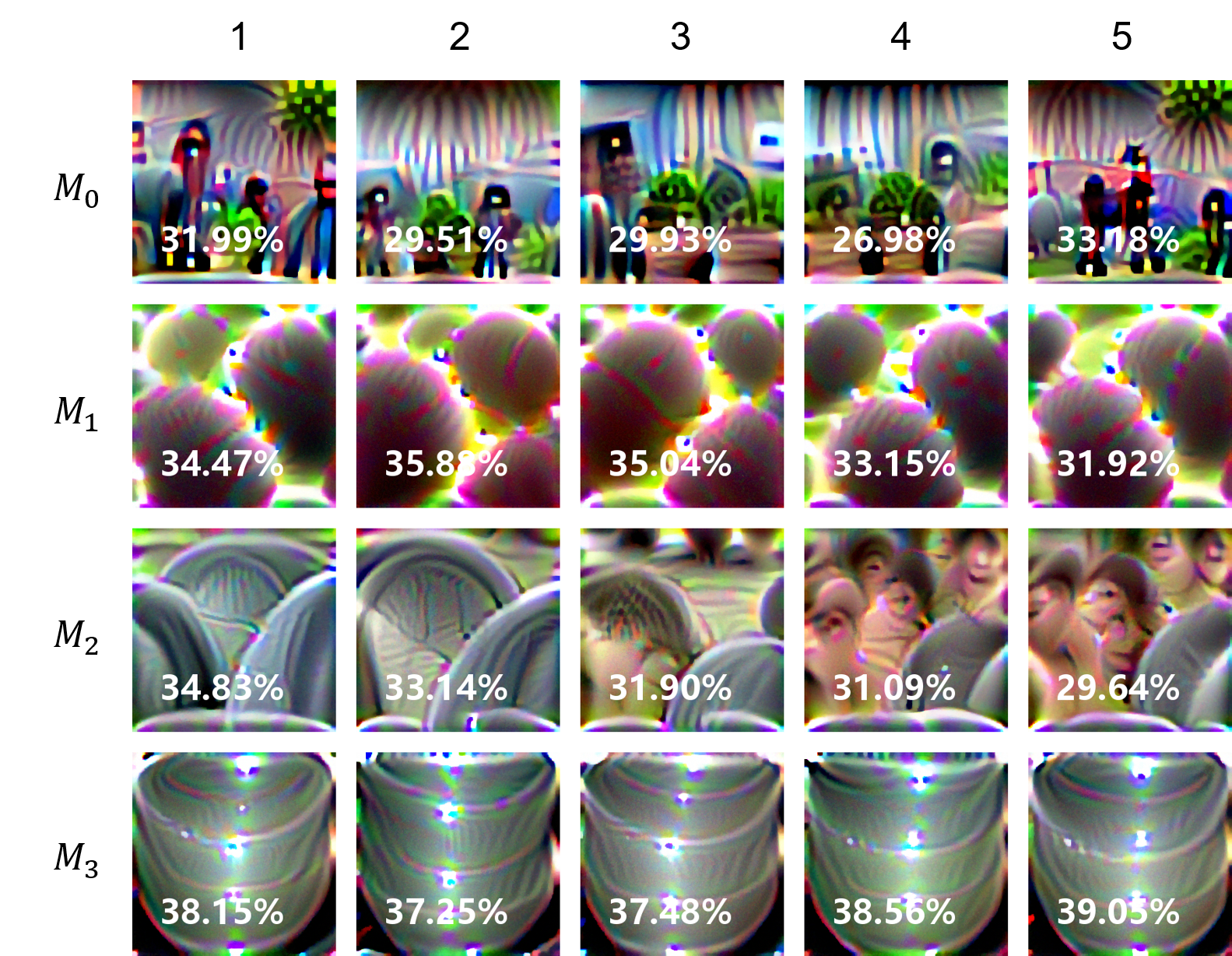}
	\caption{ Adversarial patches generated in the process of adversarial training. The five adversarial patches in each row are generated by attacking $M_0$, $M_{1}$, $M_{2}$ and $M_{3}$. The white number in the lower-left corner of each patch is the AP of $M_{0}$ when attacked by the patch.}
	\label{patches}
\end{figure}

In order to improve the training effect of Ad-YOLO, we constructed a patch dataset that containing patches with diversity and adversariality. The former ensures the generalization for Ad-YOLO, while the latter guarantees a good defense effect. To this end, we explore adversarial training for a trial.

Adversarial training \cite{goodfellow2015explaining,szegedy2014intriguing,madry2018towards} is one of the most commonly used methods to improve the robustness of the model against adversarial attacks, and it is a process of training a neural network on a mixture of clean data and adversarial examples. In general, the purpose of adversarial training is to obtain an enhanced neural network, which is the starting point from the perspective of defense. Here, we examine this method from an attack perspective.

Adversarial training based on patch attacks can be expressed as 
\begin{equation}\label{eq-adtrain}
\max_{\delta}\min_{\theta}\sum_xL_{\theta}(x,y)+\sum_{x}L_{\theta}(A(x, \delta, t), y),
\end{equation} 

The $\min_{\theta}$ can obtain adversarial training models, and $\max_{\delta}$ can generate adversarial patches based on the trained models.
In the process of alternating iterations, we generate four models $M_i, i=0,1,2,3$ and obtain multiple adversarial patches during the training of each model, some of which are shown in Figure \ref{patches}. In the figure, we can see that the patches generated from different models vary greatly, while the patches generated by the same model have relatively small differences. The AP of each patch for attacks written in the lower-left corner indicates that all patches have good attack performance.
\section{Experiments}
\subsection{Setup}
\begin{figure}[htbp]
	\centering
	\includegraphics[width=0.9\linewidth]  {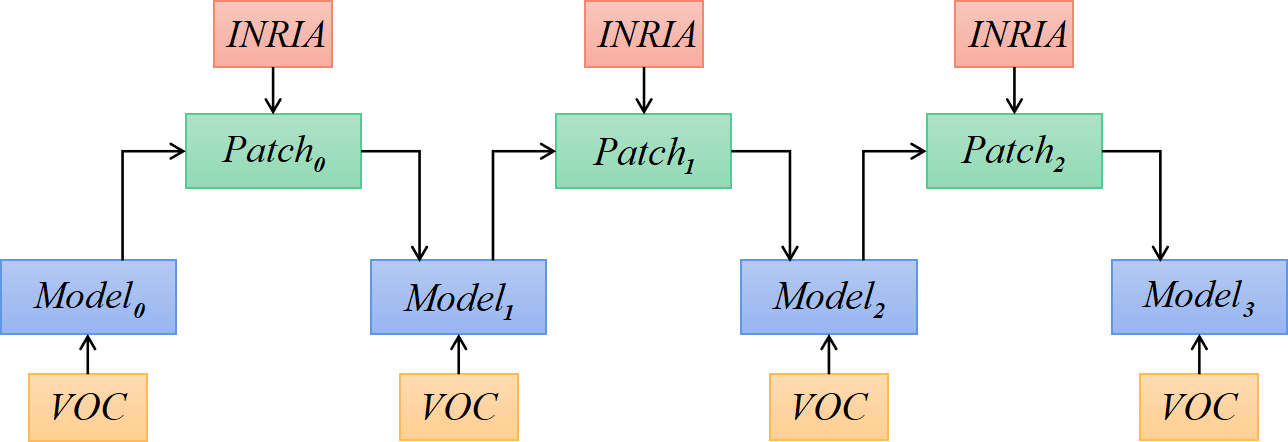}
	\caption{ Overview of the process for constructing the Inria-Patch dataset }
	\label{AT}
\end{figure}

\textbf{Dataset}
We leverage both the 2007+2012 \cite{everingham2010pascal} Pascal VOC dataset and the YOLOv2 model for the human detection task. In the construction of the Inria-Patch dataset, we first train the YOLOv2 neural network on VOC 2007+2012 training set and name it $Model_{0}$ and then, use the attack \cite{thys2019fooling} to generate adversarial patches on the Inria dataset, denoted as $patch_{0}$. Next, $patch_{0}$ are randomly added to the Inria dataset to retrain YOLOv2 with VOC 2007+2012 training set, and we obtain the new model $Model_{1}$. We carry out a total of three cycles and obtain a series of models: $M_{0}$, $M_{1}$, $M_{2}$, $M_{3}$. For the abpve four models, many more adversarial patches can be generated with the same attack method. Figure \ref{AT} shows the process for constructing the Inria-Patch dataset. For each model, five attacks with different initial values are performed, and there are a total of 20 attacks to generate adversarial patches. In the training process of each attack, 10 patches are taken out at the same step length interval, and 200 patches are obtained in total. Some patches are shown in Figure \ref{patches}. Among these patches, 160 patches are used to train Ad-YOLO, which form the dataset P0, and the remaining 40 patches form the dataset P1. For convenience, we list some simple notations, as shown in Table \ref{notation}.

\begin{table}[h]
	\centering
	\begin{tabular}{c|cc}
		\toprule
		\multirow{3}*{Dataset} & I0, I1 & Inria training set and test set  \\
		~ & P0, P1 & 160 and 40 adversarial patches  \\ 
		~ & I*-P* & Randomly add patch P* on I* \\
		\midrule
		\multirow{2}*{Model} & $M_0$ & YOLOv2 trained on Pascal VOC  \\
		~ & $M_i$ & Adversarial training models, $i=1,2,3$ \\
		\bottomrule
	\end{tabular}
	\caption{Some notations in the experiment.}
	\label{notation}
\end{table}

\textbf{Implementation details}.
In the training of the above models, the Adam optimizer is used with a learning rate of 0.01. In the training of adversarial patches, the default optimizer is Adam, and the learning rate is initialized as 0.03 and decays by a factor of 0.1 every 50 epochs.
\begin{table*}[!t]
	\centering
	{	\setlength{\tabcolsep}{1.8mm}{\begin{tabular}{cccccccccccc}
				\toprule
				{class}&aero&bike&bird&boat&bottle&bus&car&cat&chair&cow&table\\
				
				YOLOv2&83.18&80.53&73.61&66.86&51.12&76.19&83.33&81.86&56.64&68.88&73.97\\
				
				Ad-YOLO&82.31&79.36&71.05&68.82&50.19&76.40&81.44&83.50&53.42&68.13&72.82\\
				\midrule
				&dog&horse&mbike&person&plant&sheep&sofa&train&tv&mAP\\
				
				&79.31&78.19&78.52&78.70&51.20&70.59&74.06&82.31&72.02&\textbf{73.05}\\
				&78.18&76.14&77.93&77.67&48.90&68.76&73.94&85.98&71.98&\textbf{72.35}\\
				\bottomrule
	\end{tabular}}}
	\caption{Results of YOLOv2 ($Model_0$) and Ad-YOLO on VOC 2007 test set. Mean average precision and per-class average precision are shown.}
	\label{VOCmAP}
\end{table*}

All of our experiments are conducted on a computer cluster with 2 Intel Xeon E5-2650 v4 CPU cores, 128 GB RAM, and 8 NVIDIA Tesla V100 16 GB GPUs. Training each $Model_{i}$ and our Ad-YOLO requires four GPUs for four days, and training each patch requires one GPU for one day.

\subsection{Object detection}
We first conduct a series of analyses on the performance of Ad-YOLO for object detection. Table \ref{VOCmAP} demonstrates the mAP and per-class AP for YOLOv2 ($Model_{0}$) and Ad-YOLO on VOC 2007 test set. One can observe that variations in AP with each category introduced by Ad-YOLO are moderate. In the average sense, Ad-YOLO scores $72.35\%$, with a negligible decrease of $0.70\%$ from YOLOv2.

Next, we check the effectiveness of Ad-YOLO for detecting persons on I1. As shown in Figure \ref{syolo_attack}, Ad-YOLO reduces AP from $88.13\%$ for YOLOv2 to $86.91\%$, thus resulting in a $1.21\%$ decrease. In addition, patch detection for Ad-YOLO is measured on the I1-P1 dataset, and the AP reaches $83.91\%$.

Then, we seek to demonstrate the generalization of Ad-YOLO to unseen adversarial patches or persons. To this end, we evaluate a broader set of combinations of adversarial patches dataset with Inria dataset and study the effect of changing factors that are seen or unseen on the results. Table \ref{ayolo_defense_ap} shows the comparison of Ad-YOLO'AP in 4 combination cases. Specifically, Case I0-P0 is the result of the training process of Ad-YOLO and the performance is slightly better the other cases. The case I1-P0 reflects our Ad-YOLO's generalization to detect persons under seen patch attacks. In this case, Ad-YOLO achieves relatively good results, with AP reaching $78.70\%$. When facing attacks of unknown adversarial patches, in case I0-P1, Ad-YOLO can still defend the attack, and the identification accuracy of the persons reaches $78.63\%$. In the case I1-P1, the attacks and scenarios Ad-YOLO faces are both unknown, and Ad-YOLO can also successfully defend against patch attacks in this case, reaching $77.82\%$, which is the lowest value in the four cases but meets its most difficult objective conditions.

\begin{figure}[htbp]
	\begin{center}
		\includegraphics[width=0.45\linewidth]{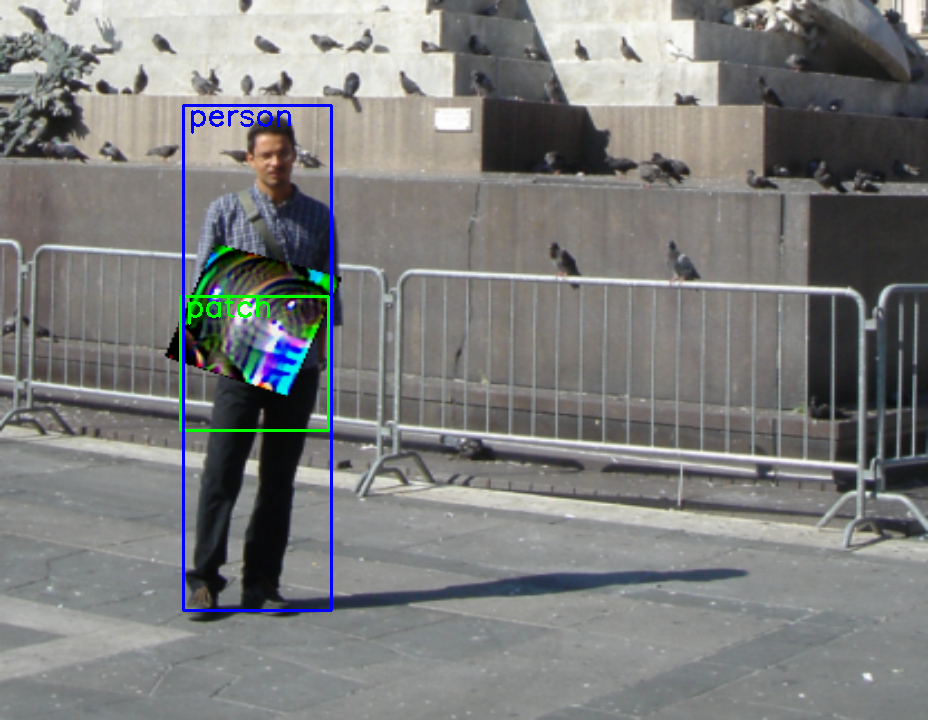}\hspace{0.1 in}
		\includegraphics[width=0.45\linewidth]{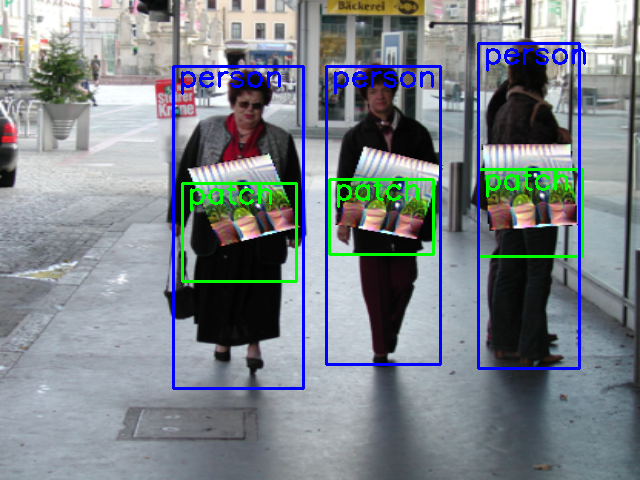}
		\caption*{(a)\hspace{1.5 in}(b)}
		\includegraphics[width=0.45\linewidth]{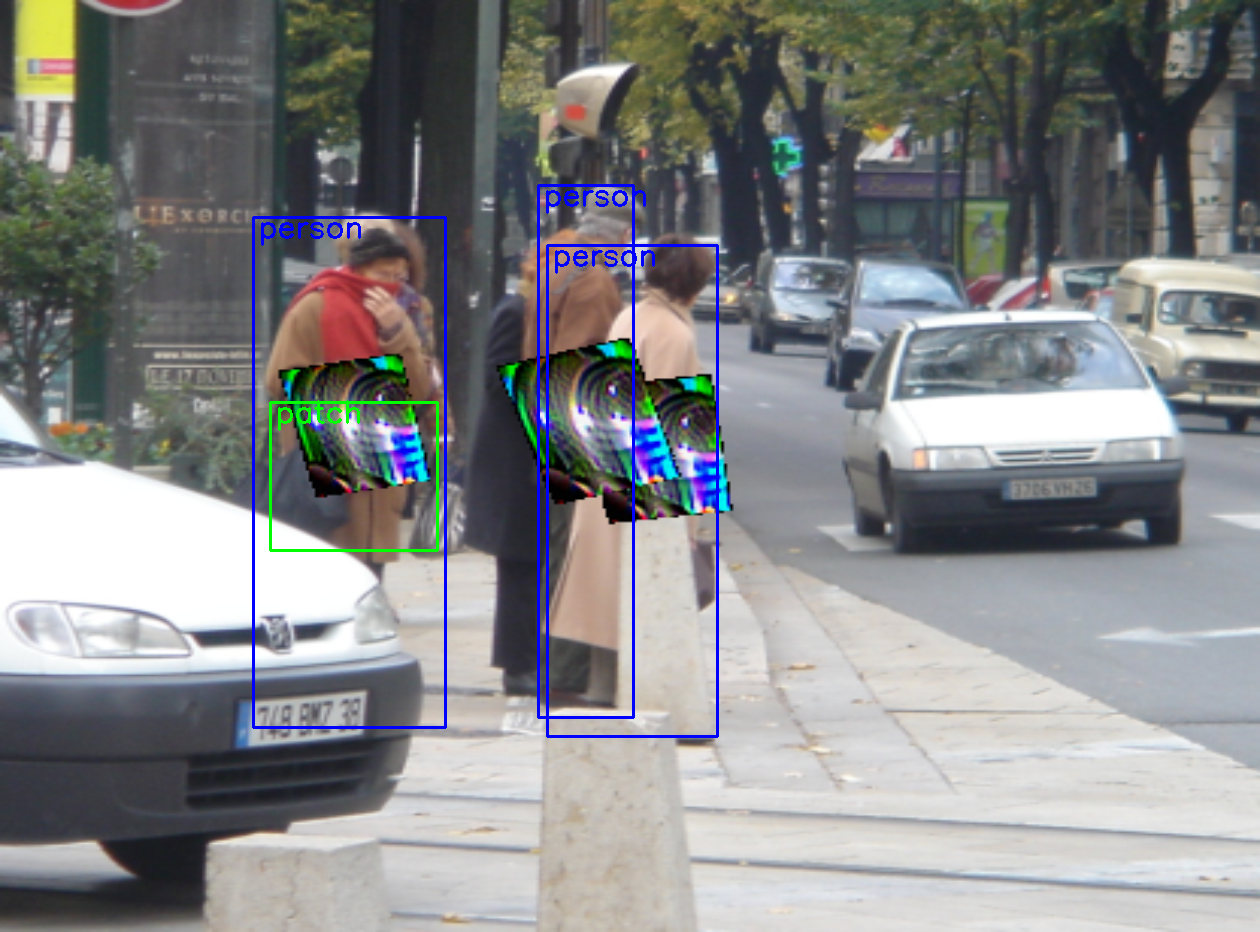}\hspace{0.1 in}
		\includegraphics[width=0.45\linewidth]{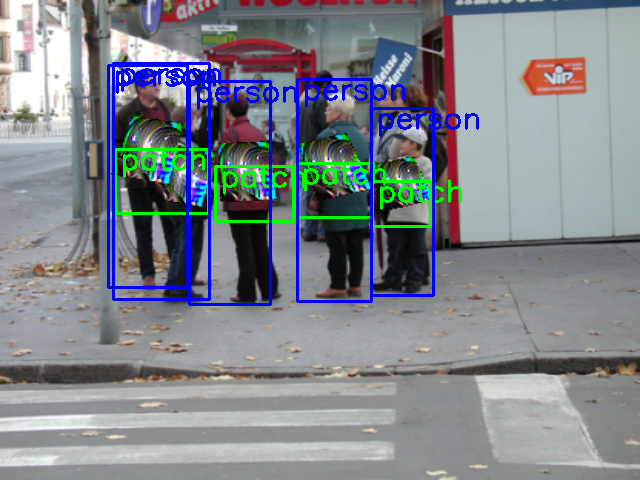}
		\caption*{(c)\hspace{1.5 in}(d)}
	\end{center}
	\caption{Display of Ad-YOLO's output in the case I1-P1. (a) single person; (b) multiple persons without overlapping; (c) and (d) multiple persons with overlapping.} \label{display}
\end{figure}

Figure \ref{display} is the display of defense in the case I1-P1. Our Ad-YOLO can simultaneously identify person and patch information under various circumstances. However, as shown in Figure \ref{display}(c), when a patch overlaps with others, Ad-YOLO may fail to detect patch information but does not affect human detection. The better results are also shown in Figure \ref{display}(b) and Figure \ref{display}(c) when the patch does not overlap in the case of one or more than one person. In Figure \ref{display}(d) shows a more complicated scenarios with more people and overlapping relationship. We can see all persons are successfully detected.

\begin{table}[h]
	\centering
	\begin{tabular}{cc}			
		\toprule
		Combination form&AP\\
		\midrule
		Training case: I0-P0&79.70\% \\
		\midrule
		Test case1: I1-P0&78.70\% \\
		\midrule
		Test case2: I0-P1&78.63\% \\
		\midrule
		Test case3: I1-P1&77.82\% \\
		\bottomrule		
	\end{tabular}
	\caption{Performance of generalization for Ad-YOLO. I0-P0 represents the training case and achieves the highest AP for detecting persons. Three test cases represent unseen persons, unseen adversarial patches and the both.}
	\label{ayolo_defense_ap}
\end{table}

To summarize, Ad-YOLO matches the property of \textbf{detectability} that a qualified defense should have. On one hand, Ad-YOLO shows a powerful strength in detecting persons on the Inria dataset without severe impact on mAP for Pascal VOC dataset. On the other hand, Ad-YOLO shows good generalization to different scenes and persons.

\subsection{Defense against patch attacks}
The second battery of experiments seek to demonstrate the defense of Ad-YOLO against white-box attacks together with physical-world attacks, and the results show that Ad-YOLO meets the requirement for \textbf{robustness}.

\subsubsection{White-box attacks}
\begin{figure}	[htbp]
	\centering
	\includegraphics[width=3 in]{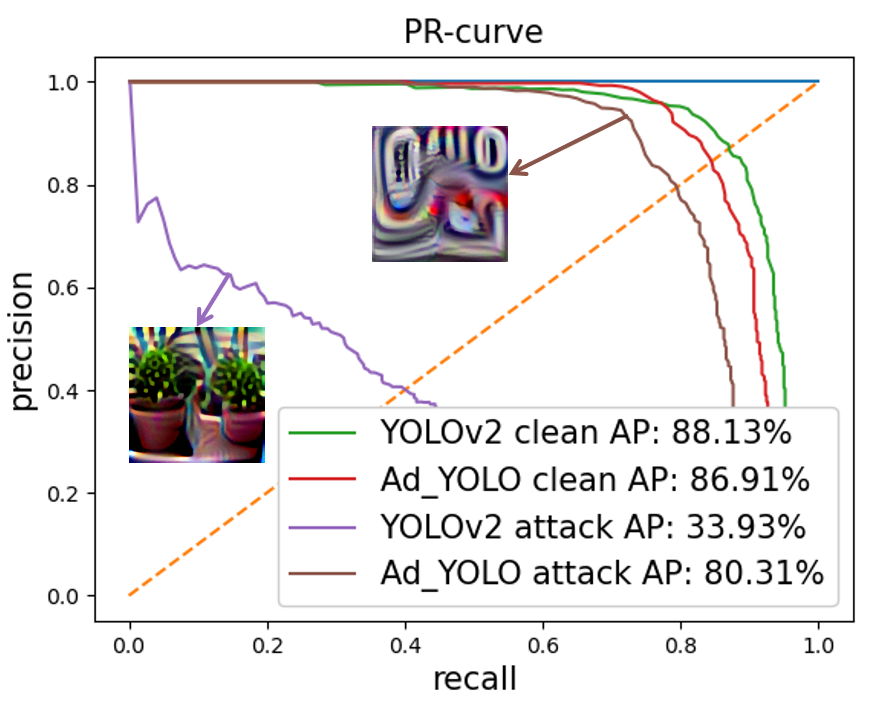}
	\caption{Performance of Ad-YOLO compared with YOLOv2 against white-box patch attack. Adversarial patches generated to attack YOLOv2 and Ad-YOLO are demonstrated with arrows.}
	\label{syolo_attack}
	\centering
\end{figure}

Here, we evaluate white-box patch attacks on Ad-YOLO. First, we leverage the same attack method \cite{thys2019fooling} to generate adversarial patches against YOLOv2 and Ad-YOLO. Then, we use generated patches to attack the two models. Figure \ref{syolo_attack} shows the performance of Ad-YOLO and YOLOv2 against white-box attacks. As we can see from the figure, two patches are considerably different due to the distinct structure of the two models, and both attacks hinder the detection of the corresponding model (YOLOv2'AP decreases from $88.13\%$ to $33.93\%$, Ad-YOLO'AP decreases from $86.91\%$ to $80.30\%$), However, Ad-YOLO shows a particularly better defense, a $\sim 46\%$ increase in AP for human detection. 

\subsubsection{Physical-World attacks}
\begin{figure*}[htbp]	
	\begin{center}
		\includegraphics[width=0.22\linewidth]{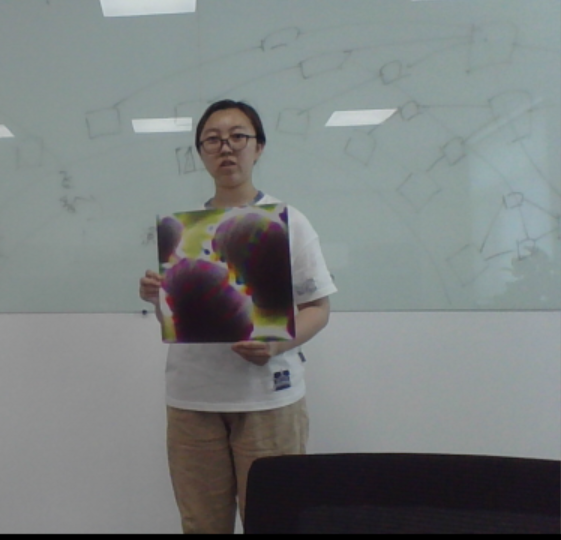}\hspace{0.05 in}
		\includegraphics[width=0.22\linewidth]{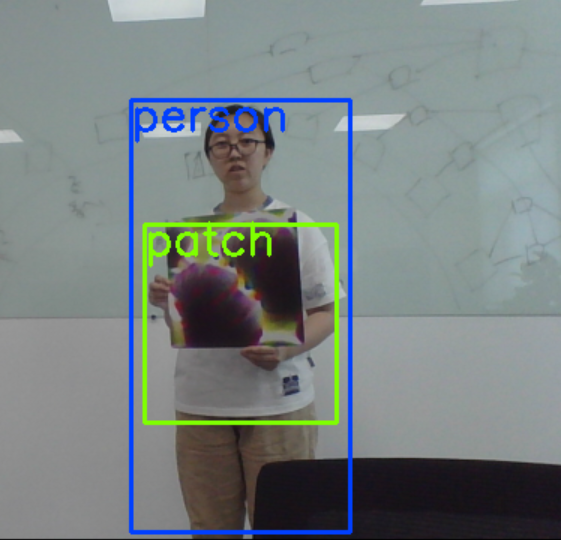}\hspace{0.05 in}
		\includegraphics[width=0.22\linewidth]{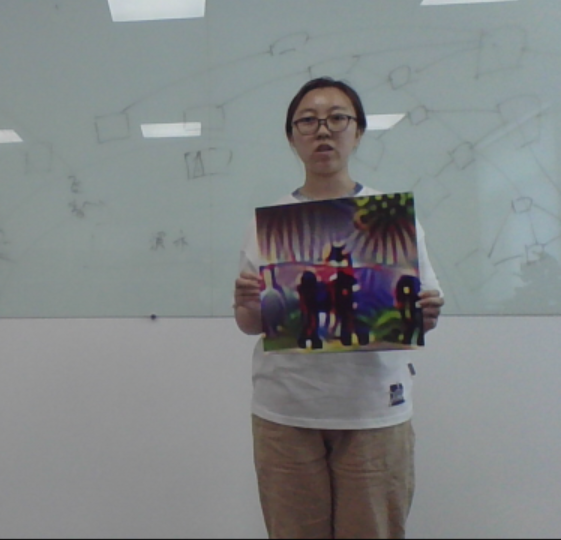}\hspace{0.05 in}
		\includegraphics[width=0.22\linewidth]{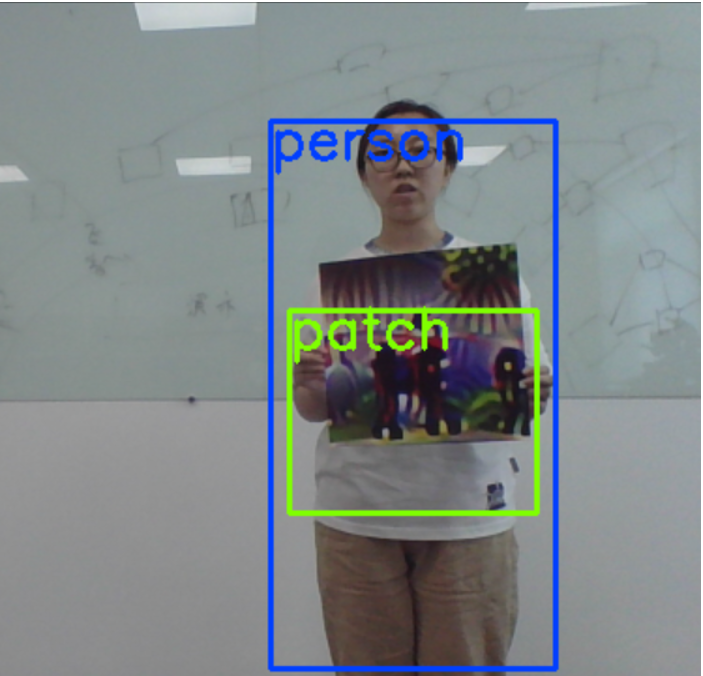}\hspace{0.05 in}
		\caption*{(a)\hspace{1.44 in}(b)\hspace{1.44 in}(c)\hspace{1.44 in}(d)}
		
		\includegraphics[width=0.22\linewidth]{./figure/black21.png}\hspace{0.05 in}
		\includegraphics[width=0.22\linewidth]{./figure/black22.png}\hspace{0.05 in}
		\includegraphics[width=0.22\linewidth]{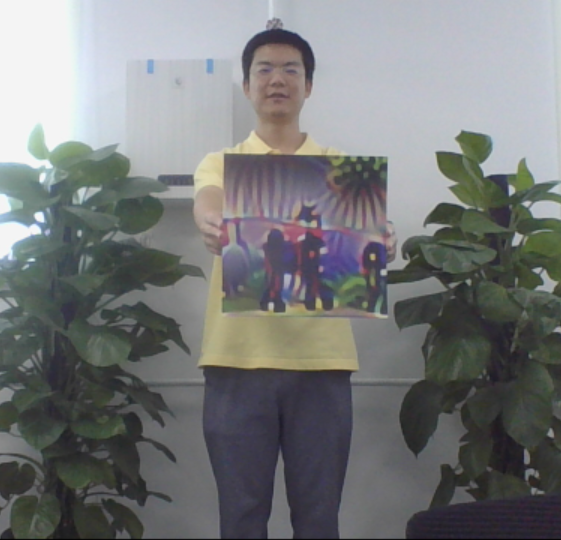}\hspace{0.05 in}
		\includegraphics[width=0.22\linewidth]{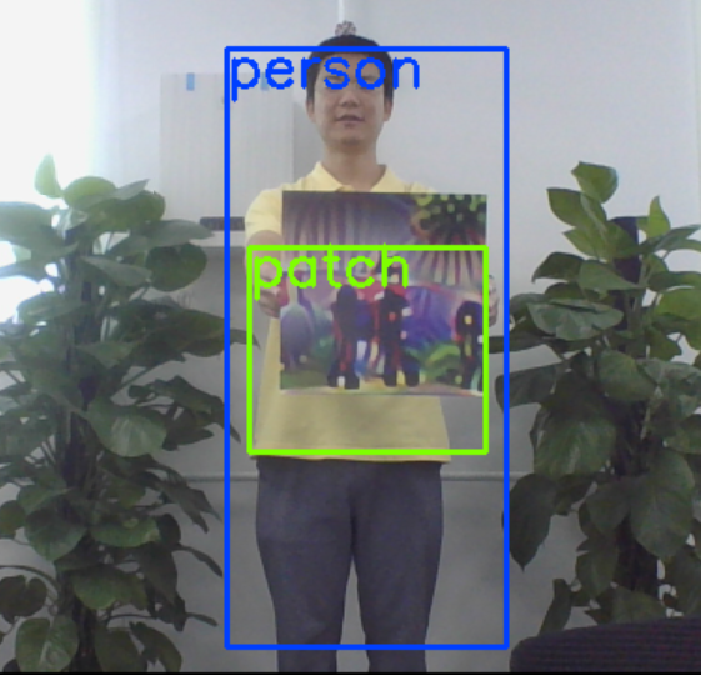}\hspace{0.05 in}
		\caption*{(e)\hspace{1.44 in}(f)\hspace{1.44 in}(g)\hspace{1.44 in}(h)}
		\caption{Display of physical-world attacks. The two rows correspond to different application scenarios. (a) (c) and (e) (g) represent the performance of YOLOv2 against patch attacks. (b) (d) and (f) (h) represent the performance of Ad-YOLO against the same adversarial patches attacks. }
		\label{display_realworld}
	\end{center}%
\end{figure*}

We evaluate Ad-YOLO against physical-world attacks. In these attacks, adversarial patches are printed and designed to be placed on people's chests. We choose different scenes to further demonstrate the defense effects. Specifically, we design 2 different scenes to compare the original detector ($Model_0$) with Ad-YOLO against physical-world attacks, and we use the identical adversarial patches randomly selected in P1.

Figure \ref{display_realworld} shows that the adversarial patches successfully attacked the original detector but failed to fool Ad-YOLO. When attacked, our Ad-YOLO could not only successfully defend against the patch attack but also recognize the patch information.

\subsection{Comparison with adversarial training models}
Adversarial training approach is one of the most widely used defense method to effectively reinforce the model's robustness to resist the adversarial attacks of white-box in classification problems, but it is rarely used in object detection problems. This part compares and shows the outstanding points of our Ad-YOLO and adversarial training method in object detection problems. In the process of developing the patch dataset, we obtain a series of adversarial train models $M_i, i=1,2,3$ and corresponding adversarial patches $Patch_i, i=0,1,2$ generated by white-box attacks. 

\begin{table}[htbp]
	\centering
	\begin{tabular}{ccc}
		\toprule
		{Model}&Clean accuracy on P0&White-box attack \\
		\midrule
		$Model_{0}$&$88.03\%$&$33.93\%$\\
		\midrule
		$Model_{1}$&$86.44\%$&$58.62\%$\\
		\midrule
		$Model_{2}$&$87.45\%$&$56.38\%$\\
		\midrule
		$Model_{3}$&$87.00\%$&$61.65\%$\\
		\midrule
		Ad-YOLO&$\textbf{86.58\%}$&$\textbf{80.31\%}$\\
		\bottomrule
	\end{tabular}
	\caption{Performance of Ad-YOLO on P0 dataset and against white-box attacks to detect persons in comparison with adversarial training models. }
	\label{adtraining}
\end{table}
Table \ref{adtraining} presents the results comparing the effectiveness of Ad-YOLO on P0 dataset and against white-box attacks to adversarial training models. In line with expectations, adversarial training models can improve the robustness to resist adversarial attack, while the clean AP decreases. However, the effect and cost of upgrading are obviously much worse than that of classification, it shows about $23\%-28\%$ increase in white-box attacks but with $1\%-2\%$ decrease in clean AP. In addition, we can see that the best performance in terms of adversarial robustness in adversarial training models is achieved by $Model_3$ when facing the white-box attack, but the result AP $61.65\%$ is still far worse than Ad-YOLO with AP $80.31\%$. Our Ad-YOLO has increased robustness by around $46\%$ but at the cost of $1.4\%$ clean AP decrease. On the basis of such a significant improvement, the drop of clean AP of Ad-YOLO is negligible compared to the original model ($Model_0$).

Ad-YOLO not only yields a significantly more robust detector, but also requires very little computing resources. As we know adversarial training methods need to alternate iterative training against attack patches and model parameters, this requires several times the computational cost of normal training and cannot be parallelized. Contrast with that, Ad-YOLO needs to be implemented to prepare the patch dataset which can be easily obtained by parallel training. Besides, compared with the normal training, the network structure and core training process of Ad-YOLO have no essential increase. Thus Ad-YOLO can performs great improvement in AP facing adversarial attacks (around $46\%$ increase), with negligible cost of clean AP ($1.4\%$ decrease) and the same computational cost as normal training (significant advantages over adversarial training).
\section{Conclusion}
At present, most defenses against patch attacks are aimed at image classification, and there are few studies on objection detection models. We focus on the human detection in object detection system and propose a novel improved defense model Ad-YOLO and construed a Inria-Patch dataset with diversity and adversariality. Through the algorithm design and numerical experiments, Ad-YOLO meets the characteristics of Timeliness, Detectability and Defense. For timeliness, We add a category (patch class) on the YOLOv2 architecture at the network structure level, which has a negligible inference increment and very little extra training cost. For Detectability, the mAP results obtained by Ad-YOLO are almost the same as those of YOLOv2 ($73.05\%$ to $72.35\%$ on VOC 2007 in Table~\ref{VOCmAP}, $88.13\%$ to $86.91\%$ on Inria in Figure~\ref{syolo_attack}). For Defense, Ad-YOLO can improve white-box defensive performance from $33.93\%$ to $80.31\%$, it is significantly better than other contrast methods. Therefore, Ad-YOLO not only meets people's needs for human detection, but also shows good defense effects against white-box attacks and physical-world attacks.
\section*{Acknowledgements}
This work was supported in part by the Innovation Foundation of Qian Xuesen Laboratory of Space Technology, and in part by Beijing Nova Program of Science and Technology under Grant Z191100001119129. 
\bibliographystyle{ieee_fullname}
\bibliography{Ad-YOLO}
\end{document}